\documentclass[letterpaper, 10 pt, conference]{ieeeconf}  

\IEEEoverridecommandlockouts                              

\usepackage{graphicx} 
\usepackage{algpseudocode}
\usepackage{algorithm}
\usepackage{caption}
\usepackage{subcaption}

\title{\LARGE \bf
Bounding Distributional Shifts in World Modeling\\ through Novelty Detection
}

\author{Eric Jing $^{1}$ and Abdeslam Boularias $^{1}$
\thanks{$^{1}$ The authors are with the Department of Computer Science, Rutgers
University, 08854 New Jersey, USA. This work is
supported by NSF awards 1846043 and 2132972. }%
}

    \newcommand{\ourMethod}{WM-VAE}

\begin{document}

\maketitle
\thispagestyle{empty}
\pagestyle{empty}

\begin{abstract}

Recent work on visual world models shows significant promise in latent state dynamics obtained from pre-trained image backbones. However, most of the current approaches are sensitive to training quality, requiring near-complete coverage of the action and state space during training to prevent divergence during inference. To make a model-based planning algorithm more robust to the quality of the learned world model, we propose in this work to use a variational autoencoder as a novelty detector to ensure that proposed action trajectories during planning do not cause the learned model to deviate from the training data distribution. To evaluate the effectiveness of this approach, a series of experiments in challenging simulated robot environments was carried out, with the proposed method incorporated into a model-predictive control policy loop extending the DINO-WM architecture. The results clearly show that the proposed method improves over state-of-the-art solutions in terms of data efficiency.
\end{abstract}

\section{INTRODUCTION}

With the increased adoption of deep learning in robotics,
{\it world models} are becoming a popular approach in model-based planning~\cite{journals/corr/abs-1803-10122,matsuo2022deep}. World models make use of pre-trained vision models to obtain latent representations of a robot's environment, replacing previous manually-crafted features~\cite{doi:10.1177/027836498600500306}. The popularization of transformers and other recent architectures made it possible to represent the state of an environment as latent features derived from structured, multimodal data. The same techniques that can generate entire video sequences from a single image frame can also find applications in planning problems in robotics, where the planner should predict how an environment evolves, conditioned on a series of actions~\cite{Boularias_Bagnell_Stentz_2014}.

However, the growing reliance on learned models in planning means that the performance of these planners is near-completely determined by the quality of the learned model. Training a reliable world model requires sufficient data to cover the various state-action cases a planner could query~\cite{levine2020offlinereinforcementlearningtutorial}. For many planners that sample random trajectories, such as MCTS and CEM, this means exhaustive coverage of all trajectories and states. With the increase in available computing power, it has become possible to achieve such coverage in controlled environments. However, in most cases, it is impractical to gather enough data for total coverage. Furthermore, in complex environments there may be unknown special cases that gathering data with a random policy may fail to cover~\cite{10.1007/s10994-022-06268-8,ShaojunIJCAI2018}. The possibility of such special cases being present means that, in practice, world models always have gaps in their knowledge.

To address this distribution shift in action trajectories with a world model between training and inference, we propose in this work to train a {\it novelty detection} model on the same dataset provided to the world model, such that every prediction by the world model is accompanied by a metric showing how far it is from the training data. The result can then influence the planner to avoid trajectories that do not lead to high-confidence predictions from the world model.

Parallel to the development of world models, novelty detection has also benefited from progress in deep learning. Where previously novelty detection relied upon statistical methods and manual feature extraction that varied considerably depending on context and type of data, current implementations take advantage of learned models such as autoencoders to directly train on data~\cite{10.1145/3205651.3208204}. This means that it is possible to perform novelty detection in a data-driven manner, in which the distribution of non-anomalous data is defined by provided training data. Although novelty detection is used primarily to analyze existing data, it can also be used to provide a confidence metric for another predictive model.

We present \ourMethod, an image-based world model with a novelty detection component that feeds into a high-level planner. Our main contributions are:
\begin{itemize}
    \item A variational autoencoder (VAE) that we use as a novelty detection component for a world model. The VAE receives as input the predicted world state and outputs a reconstruction of the state. From both the input and output, a reconstruction loss is derived during inference.
    \item A planner that uses per-action costs based on the reconstruction loss returned by the novelty detection component. Planning algorithms that randomly sample action trajectories such as CEM typically compare the end state with a goal state to derive the cost of an action trajectory. With this approach, the reconstruction loss of each predicted latent state is treated as the per-action cost for each action in a trajectory.
\end{itemize}

The proposed technique is evaluated on challenging benchmarks in simulation that involve manipulating non-rigid objects and a large number of granular objects. The results of the experiments clearly show that our proposed novelty detection and planning mechanisms significantly improve the performance of the recent world-model architecture known as {\it DINO-WM}~\cite{zhou2024dinowmworldmodelspretrained}.

   \begin{figure}[t]
      \centering
      \includegraphics[width=0.4\textwidth]{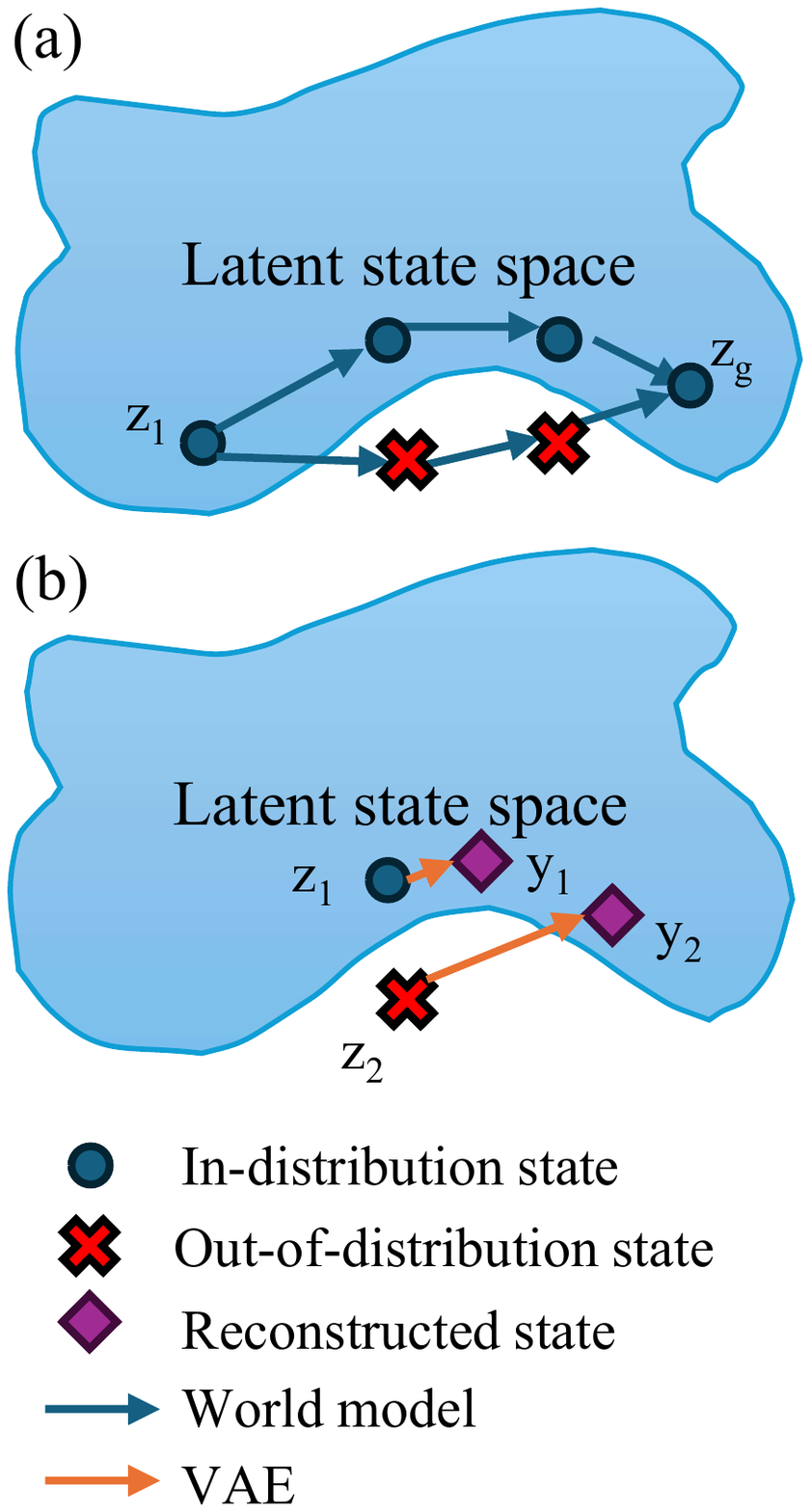}
      \caption{Depiction of potential action trajectories a planner may sample, along with predicted states in the latent space (a). Novelty detection is implemented through use of a variational autoencoder (VAE) such that out-of-distribution states incur higher reconstruction loss than in-distribution states (b).}
      \label{latentimage}
   \end{figure}

\section{RELATED WORKS}
\subsection{Deep model-based planning}
Contemporary model-based planning algorithms often use deep learning techniques to obtain latent features as  states~\cite{wang2019exploringmodelbasedplanningpolicy,hamrick2021roleplanningmodelbaseddeep}. The models are typically trained offline, taking advantage of off-board computing and existing large datasets~\cite{tian2020modelbasedvisualplanningselfsupervised}. To lessen the computational requirement for these models, it is common practice for planning algorithms working with image data to include a pretrained backbone to extract features from the input~\cite{zhou2024dinowmworldmodelspretrained}.

An issue facing model-based planning algorithms is how to ensure that generated trajectories are physically feasible~\cite{pmlr-v164-song22a}. Many algorithms use a receding horizon in the future to search for a desirable action trajectory. As predictions are generated autoregressively, errors can accumulate such that predictions in the far future become unreliable. One method to bypass this in planning is to use model-predictive control (MPC): execute the first few actions in the predicted trajectory, then re-plan starting from the new state~\cite{pmlr-v144-jain21a}. However, this does not solve the underlying issue of unreliable future predictions, which our method directly addresses.

\subsection{World modeling}
Since their inception, world models have shown promise in handling complex simulated environments~\cite{NEURIPS2018_2de5d166}. A common task for these models is to predict the physical dynamics of a given scene~\cite{doi:10.1177/02783649231169492}. Some methods explicitly model environments as a dynamical system~\cite{mao2025physicallyinterpretableworldmodels}. Others frame the problem as a form of video prediction, where sequences of images are interpolated or extrapolated~\cite{wang2024worlddreamergeneralworldmodels,chi2024evaembodiedworldmodel}. 

Because their predictions may be used as input to autoregressively predict future horizons, world models are sensitive to training and can degrade in performance during inference when predicting novel action trajectories of unseen states. Specifically, prediction errors that are introduced with each pass go on to influence further predictions, leading to a sharp decline in model performance~\cite{asadi2019combatingcompoundingerrorproblemmultistep}. A common tactic to reduce the degradation in performance over long trajectories is to introduce a frame-skip hyperparameter, in which the model is trained to take multiple actions concatenated to predict the latent state multiple steps into the future~\cite{kalyanakrishnan2021analysisframeskippingreinforcementlearning}. There are also efforts to predict entire action trajectories in one pass, using diffusion models~\cite{ding2024diffusionworldmodelfuture}. This eliminates autoregressive predictions altogether and the compounding error problem. Again, these approaches do not directly address the unreliability of future world model predictions, since they effectively enlarge the action space in order to decrease the prediction horizon.

\subsection{Novelty detection}
Also known as anomaly detection, novelty detection refers to the identification of data outside an expected distribution. Autoencoders are one of the most common ways to implement novelty detection with a learned model, particularly the variational autoencoder (VAE)~\cite{an2015variational}~\cite{nguyen2024variationalautoencoderanomalydetection}. It has many use cases, such as in intrusion detection and fraud detection~\cite{9113298}. In particular, it has even been used to detect anomalous GPS trajectories of vehicles~\cite{GAO2025112918}. Novelty detection has also been used as a confidence measure for other learned models such that a fallback policy may be substituted for a primary model whose output shows anomalous characteristics~\cite{sinha2024realtimeanomalydetectionreactive}. Our method builds upon these applications by using novelty detection in a planning loop to compute a policy.

When an autoencoder is used as part of novelty detection, the most common way to measure the score of a data sample is through reconstruction loss, where the mean-squared error (MSE) between the sample and its reconstruction is used as a similarity metric. Other reconstruction metrics can be used to calculate the score, especially for structured data like images~\cite{huijben2024enhancingreconstructionbasedoutofdistributiondetection}. Since our novelty detection model operates in the latent space, the original MSE reconstruction loss is used.

   \begin{figure*}[t]
      \centering
      \includegraphics[width=0.8\textwidth]{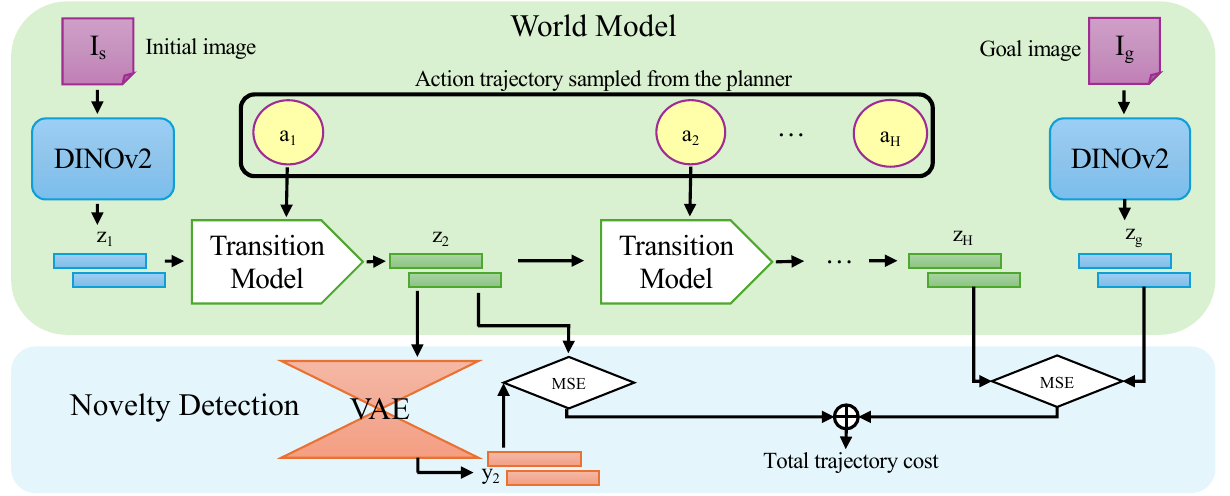}
      \caption{Overview of the world model architecture, along with the novelty detection component}
      \label{worldmodelimage}
   \end{figure*}
   
\section{APPROACH}

We present \ourMethod, a DINOv2 model-based planning algorithm that uses a VAE-based novelty detection model to address the problem of distribution shift in dynamic models. During inference, a planning algorithm uses the world model to predict future states given a sequence of actions. Each predicted future state becomes an input to the novelty detection model. From the output, a per-action cost is derived and added to the total cost at the end of the planning horizon. We present in the following the three components of \ourMethod.

\subsection{World Model}

The world model, depicted in Fig.~\ref{worldmodelimage}, is based on the DINO-WM architecture~\cite{zhou2024dinowmworldmodelspretrained}. Like DINO-WM, it consists of two modules. RGB images are first encoded into a latent representation using an encoder model. All prediction is then done in the latent space, with a transition model autoregressively generating future latent states, given an action trajectory.
Formally, given an input image $I$, let
$$
z = f(I; \theta), \eqno{(1)}
$$
 where $f$ is a function with learnable parameters $\theta$ that maps $I$ to a latent representation $z$.
Separately, given a latent representation $z_i$ and an action $a_i$, let 
$$
z_{i+1} = g(z_{i-H+1:i}, a_{i-H+1:i}; \theta), \eqno{(2)}
$$
where $g$ is a function with learnable parameters $\theta$ that maps a sequence of latent states $z_{i-H+1:i}$ and actions $a_{i-H+1:i}$ to a new latent state $z_{i+1}$. $H$ is a hyperparameter representing the size of the sliding window of past states available to $g$. 

The encoder consists of a pretrained DINOv2 network that maps image data into a sequence of patches~\cite{oquab2024dinov2learningrobustvisual}. DINOv2 has demonstrated the ability to produce rich embeddings that are capable of representing the latent state of a world model without needing to be trained specifically on image observations for a given environment~\cite{zhou2024dinowmworldmodelspretrained}. The weights of DINOv2 are frozen during training, without any fine-tuning.

The transition model uses a Vision Transformer (ViT) that operates on the sequence of patches representing the latent space to produce another sequence of patches representing the latent state corresponding to the predicted future state~\cite{dosovitskiy2021imageworth16x16words}. Low-dimensional data, such as actions and, optionally, proprioceptive data, are encoded with a linear transformation layer and appended to each patch. The final sequence of patches becomes the input to the vision transformer. Through experimentation, it was found that adding a dropout layer before the ViT increases the performance and stability of the transition model~\cite{zhou2024dinowmworldmodelspretrained}.

\subsection{Novelty Detection Model}

The novelty detection model consists of a variational autoencoder (VAE). Like all autoencoders, a VAE is trained to output a reconstruction $y$ from its input $z$, but with the additional assumption that all encountered inputs are derived from latent representations sampled from a normal distribution. Compared to a classic autoencoder, a VAE has a well-behaved latent space, which is suited to modeling the state space of a world model. Specifically, its probabilistic nature favors smooth transitions between state representations where uncertainty may be a factor~\cite{Kingma_2019}. 

In order to adapt the VAE to the patch representation of the latent space, the network is made using convolution layers, as is done with other networks operating on image data~\cite{NIPS2015_ced556cd}. The decoder half consists of transposed convolution layers, also known as deconvolution layers~\cite{5539957}. Reconstruction loss $L_r$ is defined as:

$$
L_r = MSE(y, z) \eqno{(3)}.
$$

For a given environment, the VAE is trained on latent states encoded from the same images used to train the world model. One intended consequence of this approach is that the VAE will not be able to reconstruct states that are not represented in the training data, thus reducing the ability of the model to generalize to unseen states. This is intentional, as the world model would not encounter those states during training either. Therefore, the reconstruction loss from the VAE acts as a measure of confidence in the world model for a given prediction.

\subsection{Planning Algorithm}

Since the world model does not directly predict action trajectories that lead to a desired goal state, a higher-level planning algorithm is necessary to obtain the desired action trajectory. The cross-entropy method (CEM) is used to search the action space, optimizing an action trajectory. Action trajectories are sampled from a multivariate Gaussian distribution. The transition model autoregressively predicts future latent states. The final latent state is compared to the goal state, given as an image, and a cost $L_g$ is derived. The final cost of the action trajectory $L$ is a weighted sum of $L_g$ and the reconstruction loss $L_r^{z_i}$ of every predicted latent state $z_i$.
$$
L = L_g + w\sum_i L_r^{z_i}. \eqno{(4)}
$$
$w$ represents the weight that each per-action reconstruction loss term is assigned.

Elite trajectory samples (i.e., those with lowest total cost) are used to recalculate the probability distribution, and the process repeats, as detailed in Algorithm~\ref{algorithm1}. Formally, given a collection of elite trajectory samples $A'$, the updated probability distribution is given as:
$$
a \sim \mathcal N(\bar{A'}, \widehat{\mathrm{Var}}(A')).
$$

Notably, the multivariate Gaussian distribution used in our approach is calculated from element-wise variances instead of a covariance matrix. This modification was found to be more numerically stable over multiple iterations without sacrificing the generality of the distribution~\cite{zhou2024dinowmworldmodelspretrained}.

In the context of planning, the reconstruction loss is analogous to the regularization component of many loss functions, in that it acts to moderate a predictive model and improve its ability to generalize to unseen cases. 

\begin{algorithm}[tb]
\caption{Proposed CEM Planner}
\begin{algorithmic}
    \Require Initial state $z_1$, given as an RGB image embedding
    \Require Goal state $z_g$, given as an RGB image embedding
    \State Initialize action trajectory distribution $D$ to $\mathcal N(\mathbf{0}, \mathbf{1})$
    \While{$D$ not converged}
        \State Sample a list  $A$ of $n$ trajectories from $D$ 
        \State Initialize costs $C$ of the trajectories in $A$ to $\mathbf{0}$
        \For{$k \in [1, n]$}
            \For{$i \in [1, T]$}
                \State $z_{i+1} \gets \mathrm{WorldModel}(z_i, A[k]_i)$
                \State $y_{i+1} \gets \mathrm{VAE}(z_{i+1})$
                \State $L_r^{z_{i+1}} \gets \mathrm{MSE}(y_{i+1}, z_{i+1})$
                \State $C[k] \gets C[k] + wL_r^{z_{i+1}}$
            \EndFor
            \State $L_g \gets \mathrm{MSE}(z_{T+1}, z_g)$
            \State $C[k] \gets C[k] + L_g$
        \EndFor
        \State Find elite samples $A'$ from $A$ ranked based on $C$
        \State $D \gets \mathcal N(\bar{A'}, \widehat{\mathrm{Var}}(A'))$
    \EndWhile

    \State \Return $A$
\end{algorithmic}
\label{algorithm1}
\end{algorithm}

   \begin{figure}[t]
      \centering
      \includegraphics[width=0.45\textwidth]{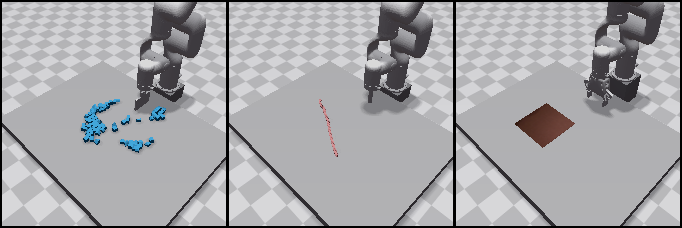}
      \caption{Environments in which WM-VAE is evaluated. From left to right: Granular, Rope, and Cloth.}
      \label{envimage}
   \end{figure}
   
   \begin{figure}[t]
      \centering
      \includegraphics[width=0.4\textwidth]{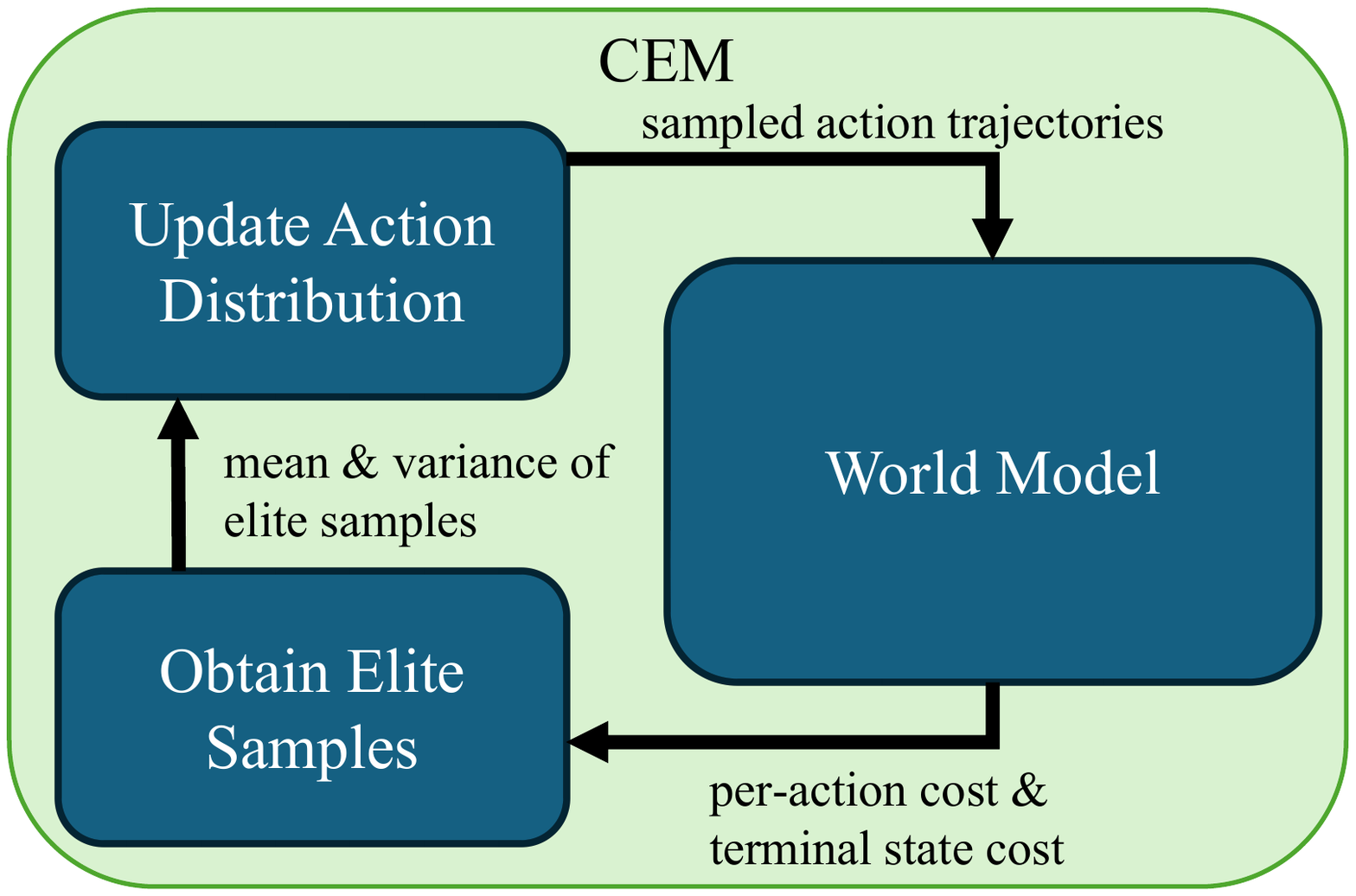}
      \caption{Overview of the CEM planner, showing the iterative optimization of trajectories in relation to the world model.}
      \label{cemimage}
   \end{figure}

\section{EXPERIMENTS}

   \begin{figure*}[t]
      \begin{subfigure}[b]{0.32\textwidth}
          \centering
          \includegraphics[width=\textwidth]{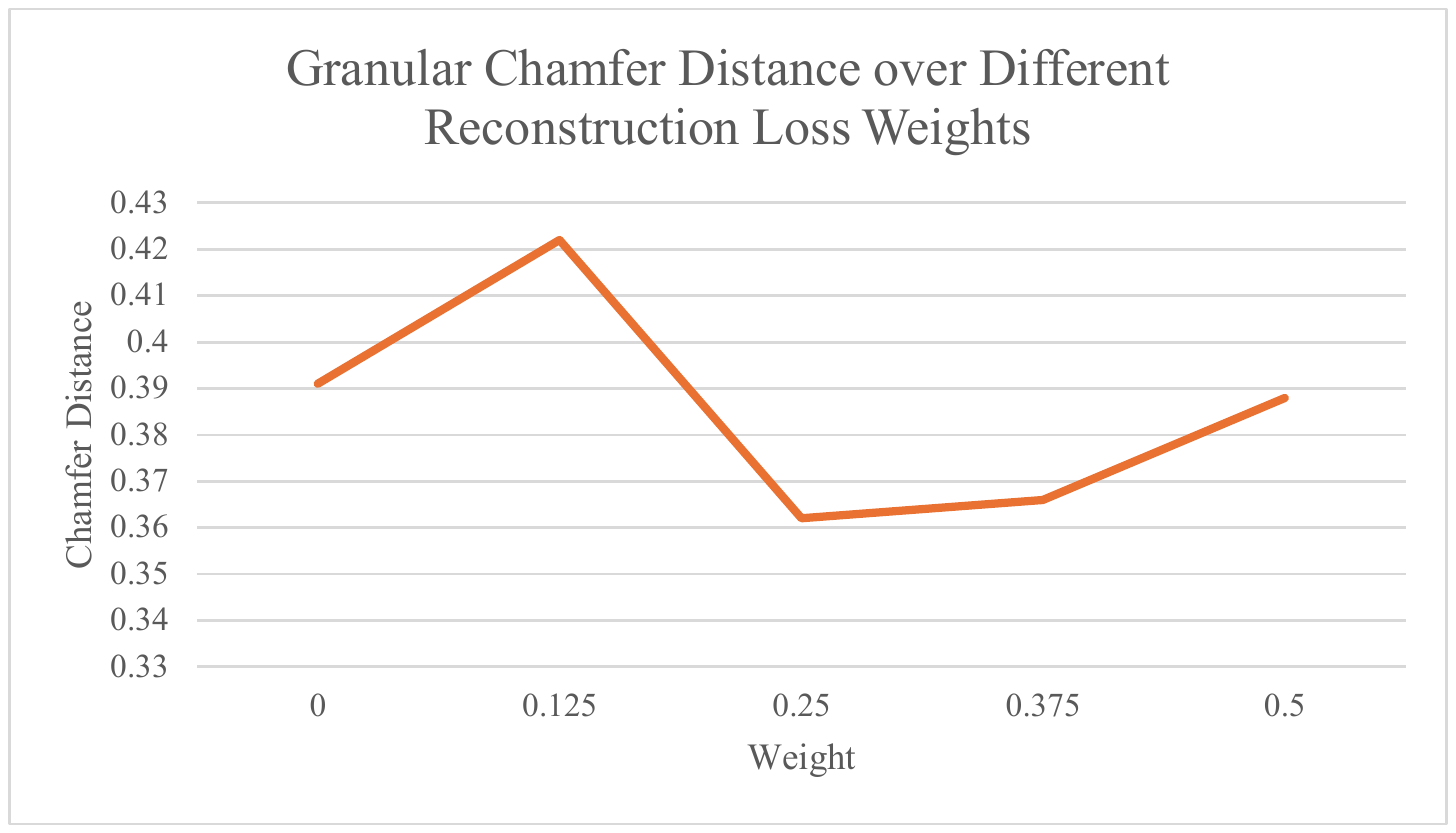}
      \end{subfigure}
      \hfill
      \begin{subfigure}[b]{0.32\textwidth}
          \centering
          \includegraphics[width=\textwidth]{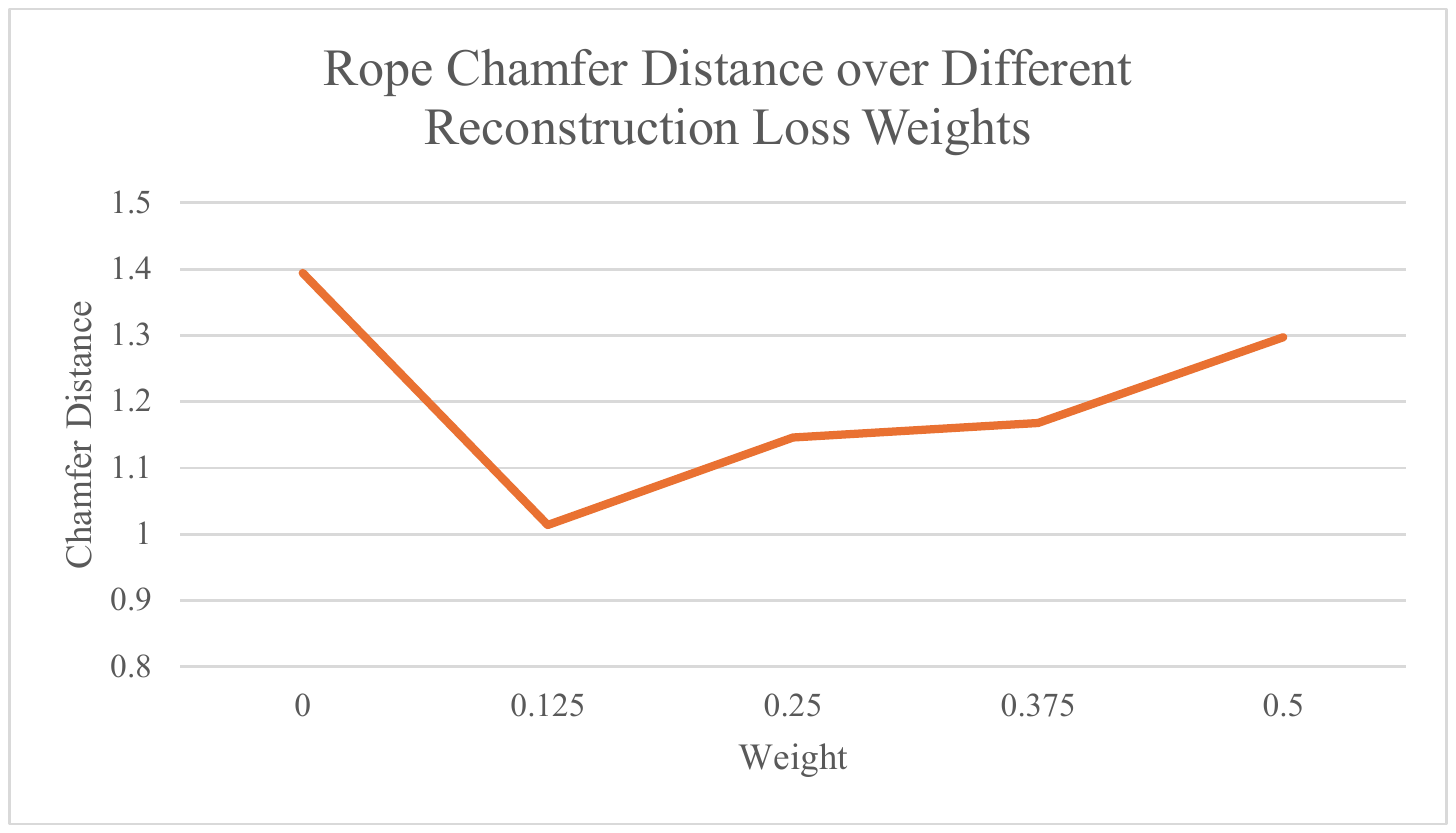}
      \end{subfigure}
      \hfill
      \begin{subfigure}[b]{0.32\textwidth}
          \centering
          \includegraphics[width=\textwidth]{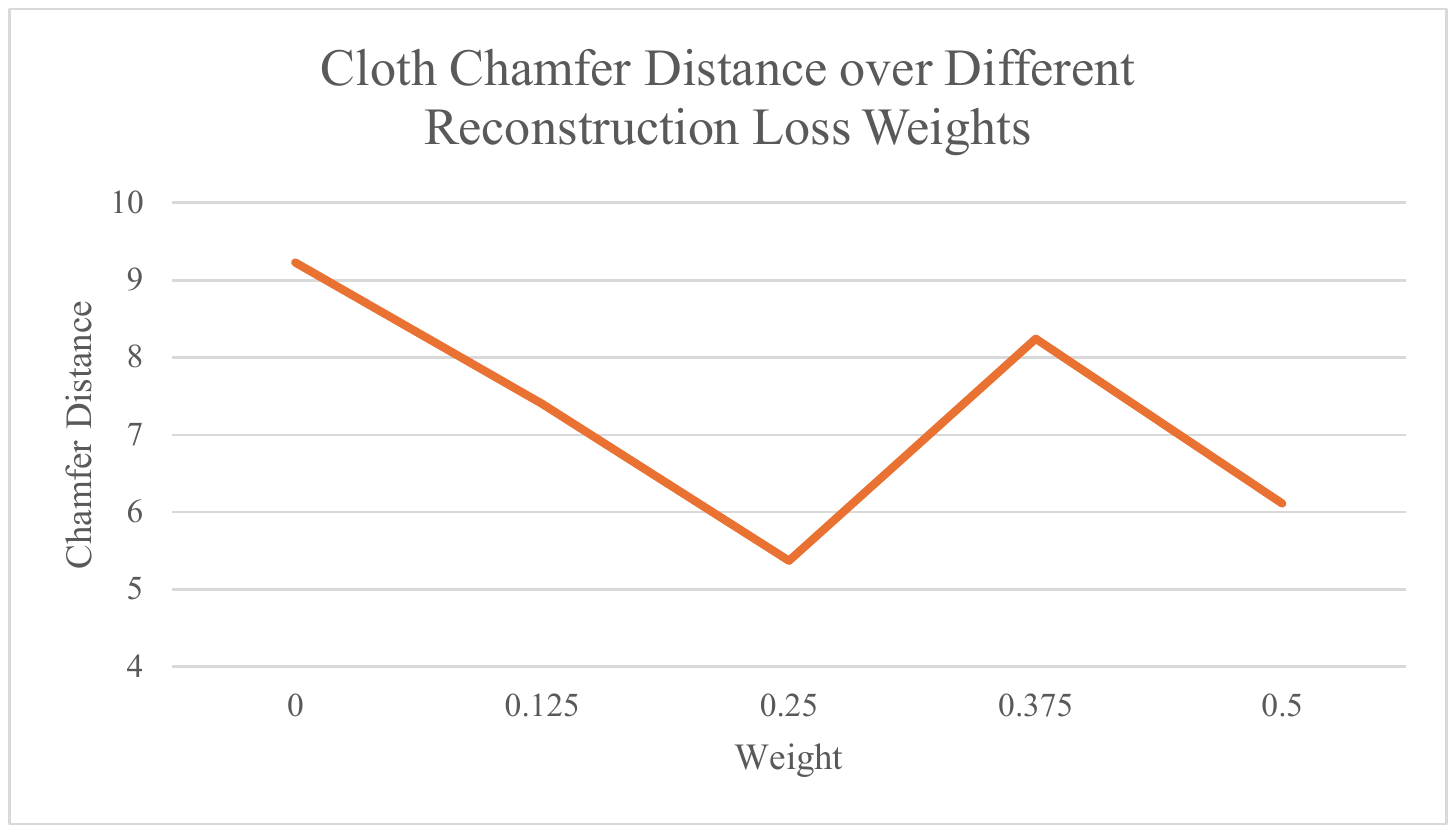}
      \end{subfigure}
      \caption{Graphs showing how Chamfer distance changes as reconstruction loss weight increases for various environments}
      \label{chart}
   \end{figure*}

   \begin{figure*}[t]
      \centering
      \begin{subfigure}[b]{0.85\textwidth}
          \centering
          \includegraphics[width=\textwidth]{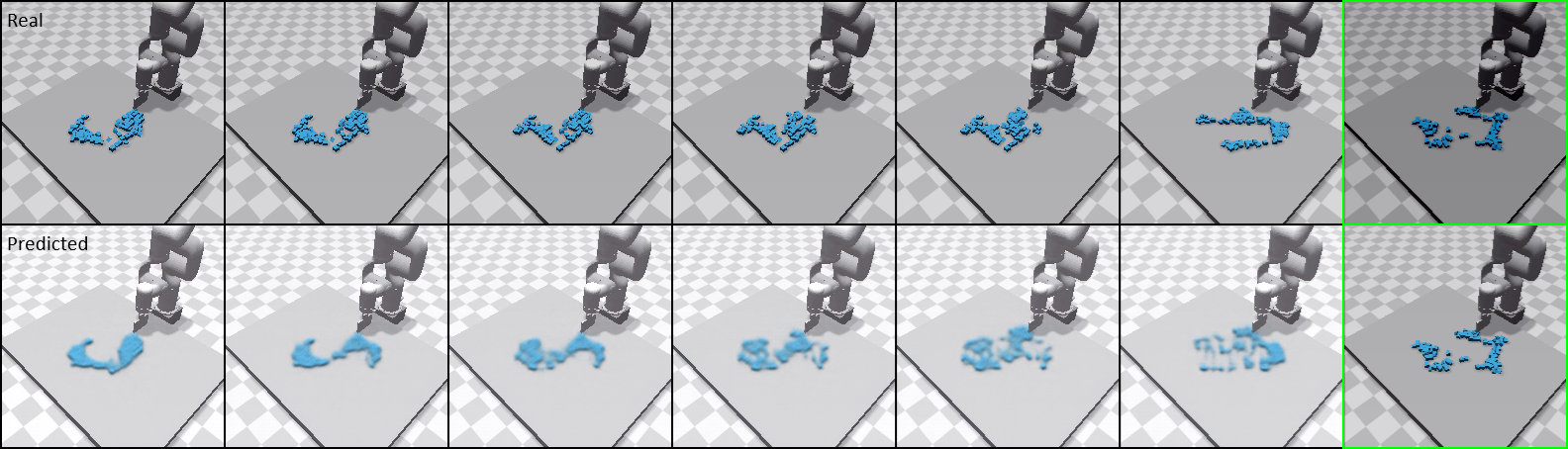}
          \caption{Granular - without reconstruction loss}
      \end{subfigure}
      \begin{subfigure}[b]{0.85\textwidth}
          \centering
          \includegraphics[width=\textwidth]{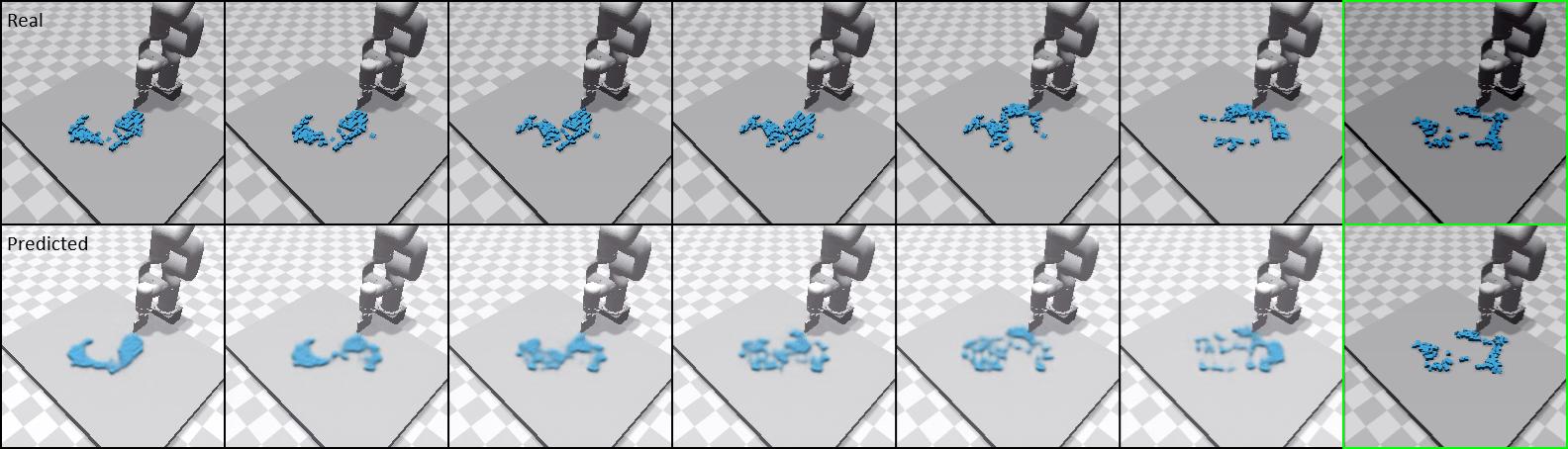}
          \caption{Granular - with reconstruction loss}
      \end{subfigure}
      \begin{subfigure}[b]{0.85\textwidth}
          \centering
          \includegraphics[width=\textwidth]{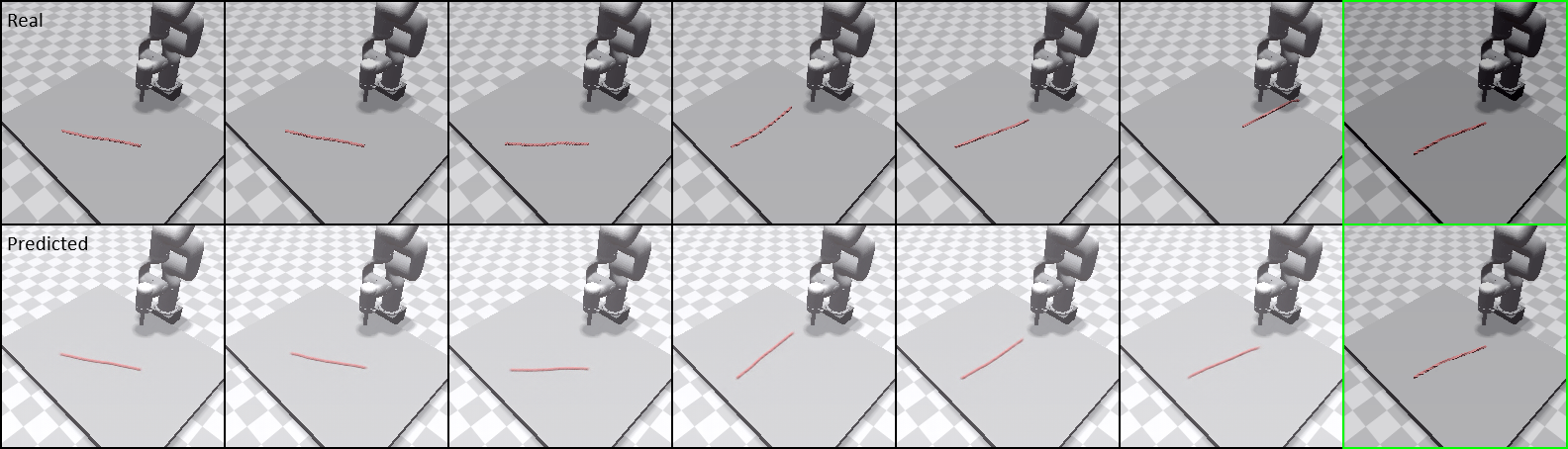}
          \caption{Rope - without reconstruction loss}
      \end{subfigure}
      \begin{subfigure}[b]{0.85\textwidth}
          \centering
          \includegraphics[width=\textwidth]{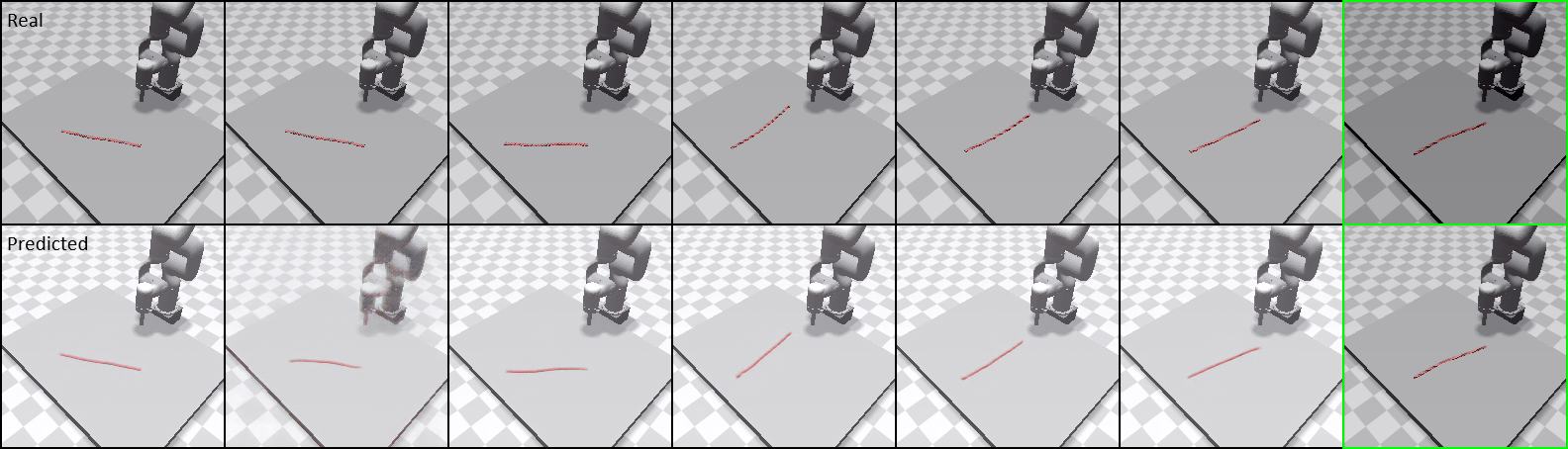}
          \caption{Rope - with reconstruction loss}
      \end{subfigure}

      \caption{Rollout images comparing actual with predicted states in various environments. For each example, the top row contains the actual observed states, and the bottom row the predicted states visualized by the decoder. The last column of each row contains the goal state.}
      \label{qualitative}
      \vspace{-0.6cm}
   \end{figure*}

\subsection{Setup}

To evaluate the performance of the proposed approach, we carried out several experiments in the following simulated environments using NVIDIA FleX.

\begin{enumerate}
    \item {\it Granular} - an environment with a collection of small particles that a robot arm pushes.
    \item {\it Rope} - an environment with a length of rope that a robot arm pushes.
    \item {\it Cloth} - an environment with a square of cloth that a robot arm grabs and places.
\end{enumerate}

An action is a planar motion defined by the starting position of the robot arm end-effector and the ending position of the robot arm end effector. In all scenarios, the end-effector positions are two-dimensional, as each motion is performed in a planar workspace. An observation is a rendered RGB image of the environment. No proprioceptive data is used from the simulated environments.
The length of the sliding window $H$ is set to 1. This means that the world model receives only the last frame to predict the next frame.

Prior works on world models have found it beneficial to introduce a frame-skip hyperparameter $F$, in which $F$ actions are concatenated together to be treated as a single action and used by the world model to predict $F$ steps into the future in one pass~\cite{kalyanakrishnan2021analysisframeskippingreinforcementlearning}. This is especially effective in environments where observations change little between frames. Given the action spaces of the FleX environments consist of whole motions rather than incremental delta commands, no frame-skip is used in this setup, which is equivalent to $F=1$.

\subsection{Training}

For each environment, the entire model is trained on actions and observations generated from a random policy. Compared to prior works, the training dataset is smaller in order to study the influence of novelty detection on an imperfect world model.

\begin{enumerate}
    \item Granular - 100 training trajectories with 20 frames
    \item Rope - 1000 training trajectories with 20 frames
    \item Cloth - 100 training trajectories with 5 frames
\end{enumerate}

\begin{table}[b]
\caption{Comparison of Euclidean and Chamfer distance between our proposed novelty detection technique WM-VAE and the state-of-the-art DINO-WM~\cite{zhou2024dinowmworldmodelspretrained} 
}
\label{table_example}
\begin{center}
\begin{tabular}{|ccccc|}
\hline
\textbf{Method} & \textbf{Granular} CD & \textbf{Rope} CD & \textbf{Cloth} CD\\
\hline
DINO-WM~\cite{zhou2024dinowmworldmodelspretrained} & 0.391 & 1.394 & 9.228\\
\hline
WM-VAE (ours) & \textbf{0.362} & \textbf{1.014} & \textbf{5.372}\\
\hline
\end{tabular}
\end{center}
\end{table}

\begin{table}[tb]
\caption{Comparison of Chamfer distance across different image encoders}
\label{table_example_backbone}
\begin{center}
\begin{tabular}{|cccc|}
\hline
\textbf{Backbone} & \textbf{Granular} CD & \textbf{Rope} CD & \textbf{Cloth} CD\\
\hline
WM-VAE with ResNet & 0.660 & 1.403 & 7.393\\
\hline
WM-VAE with DINOv2 & \textbf{0.362} & \textbf{1.014} & \textbf{5.372}\\
\hline
\end{tabular}
\end{center}
\end{table}

\begin{table}[tb]
\caption{Comparison of Chamfer distance across different reconstruction loss weights}
\label{table_example_weights}
\begin{center}
\begin{tabular}{|cccc|}
\hline
\textbf{Weight} & \textbf{Granular} CD & \textbf{Rope} CD & \textbf{Cloth} CD\\
\hline
0 &0.391 &1.394 &9.228\\
\hline
0.125 &0.422 &\textbf{1.014} &7.401\\
\hline
0.25 &\textbf{0.362} &1.146 &\textbf{5.372}\\
\hline
0.375 &0.366 &1.168 &8.246\\
\hline
0.5 &0.388 &1.297 &6.109\\
\hline
\end{tabular}
\end{center}
\end{table}

\subsection{Evaluation}

\subsubsection{Quantitative Results}
To measure the effectiveness of the novelty detection model, rollouts are performed with and without the reconstruction loss derived from the predicted latent states. The Chamfer Distance (CD) between the goal state and the end state is used as a metric, as it is commonly used to represent the distance between two point clouds. In the case of the simulated environments, the point clouds are composed of the particles that make up the scene in the FleX environment. Each method is performed with 5 different scenes. The Chamfer distances are then averaged across the scenes to obtain the final score. Table~\ref{table_example} displays the results, which suggest that the addition of the novelty detection component improves the performance of the planner. 

In addition to directly comparing the performance of DINO-WM with and without the per-state reconstruction loss, several ablation experiments are performed. Table~\ref{table_example_backbone} compares the performance of the world model when the backbone network used in the image encoder is replaced with a pretrained ResNet backbone. To adapt the novelty detection model to the ResNet embeddings, the VAE is modified with fully-connected layers replacing the convolutional layers. As DINOv2 has demonstrated state-of-the-art performance as a general-purpose image encoder, it is expected that it is more effective at encoding images into a latent representation than prior methods. 

Another experiment showcases the effect of changing the weight hyperparameter that determines how much the novelty detection model influences the trajectory cost. Table~\ref{table_example_weights} is an example of how past a certain point, increasing the novelty weight reduces the performance of the planning algorithm. We speculate that as the reconstruction loss term dominates the trajectory loss, the goal signal becomes less influential, causing the CEM planner to behave erratically and not converge toward the goal state. Another possibility is that the optimal weight hyperparameter for a given trained world model may differ depending on the overall performance of the world model with respect to the novelty detection model. A world model whose predicted latent states regularly produce high reconstruction losses may require a smaller reconstruction loss weight than a world model whose predicted latent states have low reconstruction loss.

\subsubsection{Qualitative Results}
To provide qualitative results and aid in debugging, a decoder network is trained separately to convert latent states back into images~\cite{zhou2024dinowmworldmodelspretrained}. The decoder is based on that of VQ-VAE, and it is trained by reconstructing images encoded by the pretrained DINOv2 encoder~\cite{oord2018neuraldiscreterepresentationlearning}. 

Qualitatively, the predicted latent states produced by the output action trajectories in which reconstruction loss is applied are closer to actual observation during rollout than the predictions in which no reconstruction loss is applied. This is expected, as the additional reconstruction loss cost term penalizes the planner for straying far from the dataset on which the novelty detection model is trained. Since the world model and novelty detection model are trained on the same dataset, the latent states that are poorly reconstructed by the novelty detection model are likely to be out-of-distribution in the context of the world model.

\section{CONCLUSION}

Novelty detection offers significant benefits for planning algorithms that use world models. By imposing costs for straying from in-distribution states, planners can avoid trajectories that are not adequately covered by training data. Thus, such an approach mitigates the effects of an imperfect world model on planning, as demonstrated by our simulation experiments on challenging robot manipulation problems. One limitation of this approach is that it biases against unseen states in favor of reliability. Further exploration in this direction may involve investigating the impact of world model quality on planning with novelty detection.





\section*{ACKNOWLEDGMENT}

We would like to thank Gaoyue Zhou for clarifying the implementation details of DINO-WM, as well as for providing guidance in reproducing experimental results.


\bibliographystyle{bibtex/bst/IEEEtran}
\bibliography{cited}

\begin{thebibliography}{10}
\providecommand{\url}[1]{#1}
\csname url@rmstyle\endcsname
\providecommand{\newblock}{\relax}
\providecommand{\bibinfo}[2]{#2}
\providecommand\BIBentrySTDinterwordspacing{\spaceskip=0pt\relax}
\providecommand\BIBentryALTinterwordstretchfactor{4}
\providecommand\BIBentryALTinterwordspacing{\spaceskip=\fontdimen2\font plus
\BIBentryALTinterwordstretchfactor\fontdimen3\font minus \fontdimen4\font\relax}
\providecommand\BIBforeignlanguage[2]{{%
\expandafter\ifx\csname l@#1\endcsname\relax
\typeout{** WARNING: IEEEtran.bst: No hyphenation pattern has been}%
\typeout{** loaded for the language `#1'. Using the pattern for}%
\typeout{** the default language instead.}%
\else
\language=\csname l@#1\endcsname
\fi
#2}}

\bibitem{journals/corr/abs-1803-10122}
\BIBentryALTinterwordspacing
D.~Ha and J.~Schmidhuber, ``World models.'' \emph{CoRR}, vol. abs/1803.10122, 2018. [Online]. Available: \url{http://dblp.uni-trier.de/db/journals/corr/corr1803.html\#abs-1803-10122}
\BIBentrySTDinterwordspacing

\bibitem{matsuo2022deep}
Y.~Matsuo, Y.~LeCun, M.~Sahani, D.~Precup, D.~Silver, M.~Sugiyama, E.~Uchibe, and J.~Morimoto, ``Deep learning, reinforcement learning, and world models,'' \emph{Neural Networks}, vol. 152, pp. 267--275, 2022.

\bibitem{doi:10.1177/027836498600500306}
\BIBentryALTinterwordspacing
C.~G. Atkeson, C.~H. An, and J.~M. Hollerbach, ``Estimation of inertial parameters of manipulator loads and links,'' \emph{The International Journal of Robotics Research}, vol.~5, no.~3, pp. 101--119, 1986. [Online]. Available: \url{https://doi.org/10.1177/027836498600500306}
\BIBentrySTDinterwordspacing

\bibitem{Boularias_Bagnell_Stentz_2014}
A.~Boularias, J.~Bagnell, and A.~Stentz, ``Efficient optimization for autonomous robotic manipulation of natural objects,'' \emph{Proceedings of the AAAI Conference}, vol.~28, no.~1, Jun. 2014.

\bibitem{levine2020offlinereinforcementlearningtutorial}
\BIBentryALTinterwordspacing
S.~Levine, A.~Kumar, G.~Tucker, and J.~Fu, ``Offline reinforcement learning: Tutorial, review, and perspectives on open problems,'' 2020. [Online]. Available: \url{https://arxiv.org/abs/2005.01643}
\BIBentrySTDinterwordspacing

\bibitem{10.1007/s10994-022-06268-8}
\BIBentryALTinterwordspacing
K.~Ghosh, C.~Bellinger, R.~Corizzo, P.~Branco, B.~Krawczyk, and N.~Japkowicz, ``The class imbalance problem in deep learning,'' \emph{Mach. Learn.}, vol. 113, no.~7, p. 4845–4901, Dec. 2022. [Online]. Available: \url{https://doi.org/10.1007/s10994-022-06268-8}
\BIBentrySTDinterwordspacing

\bibitem{ShaojunIJCAI2018}
S.~Zhu, A.~Kimmel, K.~E. Bekris, and A.~Boularias, ``Fast model identification via physics engines for improved policy search,'' in \emph{Proceedings of the 27th International Joint Conference on Artificial Intelligence (IJCAI), Stockholm, Sweden}, 2018.

\bibitem{10.1145/3205651.3208204}
\BIBentryALTinterwordspacing
V.~Fehst, H.~C. La, T.-D. Nghiem, B.~E. Mayer, P.~Englert, and K.-H. Fiebig, ``Automatic vs. manual feature engineering for anomaly detection of drinking-water quality,'' in \emph{Proceedings of the Genetic and Evolutionary Computation Conference Companion}, ser. GECCO '18.\hskip 1em plus 0.5em minus 0.4em\relax New York, NY, USA: Association for Computing Machinery, 2018, p. 5–6. [Online]. Available: \url{https://doi.org/10.1145/3205651.3208204}
\BIBentrySTDinterwordspacing

\bibitem{zhou2024dinowmworldmodelspretrained}
\BIBentryALTinterwordspacing
G.~Zhou, H.~Pan, Y.~LeCun, and L.~Pinto, ``Dino-wm: World models on pre-trained visual features enable zero-shot planning,'' 2024. [Online]. Available: \url{https://arxiv.org/abs/2411.04983}
\BIBentrySTDinterwordspacing

\bibitem{wang2019exploringmodelbasedplanningpolicy}
\BIBentryALTinterwordspacing
T.~Wang and J.~Ba, ``Exploring model-based planning with policy networks,'' 2019. [Online]. Available: \url{https://arxiv.org/abs/1906.08649}
\BIBentrySTDinterwordspacing

\bibitem{hamrick2021roleplanningmodelbaseddeep}
\BIBentryALTinterwordspacing
J.~B. Hamrick, A.~L. Friesen, F.~Behbahani, A.~Guez, F.~Viola, S.~Witherspoon, T.~Anthony, L.~Buesing, P.~Veličković, and T.~Weber, ``On the role of planning in model-based deep reinforcement learning,'' 2021. [Online]. Available: \url{https://arxiv.org/abs/2011.04021}
\BIBentrySTDinterwordspacing

\bibitem{tian2020modelbasedvisualplanningselfsupervised}
\BIBentryALTinterwordspacing
S.~Tian, S.~Nair, F.~Ebert, S.~Dasari, B.~Eysenbach, C.~Finn, and S.~Levine, ``Model-based visual planning with self-supervised functional distances,'' 2020. [Online]. Available: \url{https://arxiv.org/abs/2012.15373}
\BIBentrySTDinterwordspacing

\bibitem{pmlr-v164-song22a}
\BIBentryALTinterwordspacing
H.~Song, D.~Luan, W.~Ding, M.~Y. Wang, and Q.~Chen, ``Learning to predict vehicle trajectories with model-based planning,'' in \emph{Proceedings of the 5th Conference on Robot Learning}, ser. Proceedings of Machine Learning Research, A.~Faust, D.~Hsu, and G.~Neumann, Eds., vol. 164.\hskip 1em plus 0.5em minus 0.4em\relax PMLR, 08--11 Nov 2022, pp. 1035--1045. [Online]. Available: \url{https://proceedings.mlr.press/v164/song22a.html}
\BIBentrySTDinterwordspacing

\bibitem{pmlr-v144-jain21a}
\BIBentryALTinterwordspacing
A.~Jain, L.~Chan, D.~S. Brown, and A.~D. Dragan, ``Optimal cost design for model predictive control,'' in \emph{Proceedings of the 3rd Conference on Learning for Dynamics and Control}, ser. Proceedings of Machine Learning Research, A.~Jadbabaie, J.~Lygeros, G.~J. Pappas, P.~A.~Parrilo, B.~Recht, C.~J. Tomlin, and M.~N. Zeilinger, Eds., vol. 144.\hskip 1em plus 0.5em minus 0.4em\relax PMLR, 07 -- 08 June 2021, pp. 1205--1217. [Online]. Available: \url{https://proceedings.mlr.press/v144/jain21a.html}
\BIBentrySTDinterwordspacing

\bibitem{NEURIPS2018_2de5d166}
D.~Ha and J.~Schmidhuber, ``Recurrent world models facilitate policy evolution,'' in \emph{Advances in Neural Information Processing Systems}, S.~Bengio, H.~Wallach, H.~Larochelle, K.~Grauman, N.~Cesa-Bianchi, and R.~Garnett, Eds., vol.~31.\hskip 1em plus 0.5em minus 0.4em\relax Curran Associates, Inc., 2018.

\bibitem{doi:10.1177/02783649231169492}
\BIBentryALTinterwordspacing
M.~Lutter and J.~Peters, ``Combining physics and deep learning to learn continuous-time dynamics models,'' \emph{The International Journal of Robotics Research}, vol.~42, no.~3, pp. 83--107, 2023. [Online]. Available: \url{https://doi.org/10.1177/02783649231169492}
\BIBentrySTDinterwordspacing

\bibitem{mao2025physicallyinterpretableworldmodels}
\BIBentryALTinterwordspacing
Z.~Mao and I.~Ruchkin, ``Towards physically interpretable world models: Meaningful weakly supervised representations for visual trajectory prediction,'' 2025. [Online]. Available: \url{https://arxiv.org/abs/2412.12870}
\BIBentrySTDinterwordspacing

\bibitem{wang2024worlddreamergeneralworldmodels}
\BIBentryALTinterwordspacing
X.~Wang, Z.~Zhu, G.~Huang, B.~Wang, X.~Chen, and J.~Lu, ``Worlddreamer: Towards general world models for video generation via predicting masked tokens,'' 2024. [Online]. Available: \url{https://arxiv.org/abs/2401.09985}
\BIBentrySTDinterwordspacing

\bibitem{chi2024evaembodiedworldmodel}
\BIBentryALTinterwordspacing
X.~Chi, H.~Zhang, C.-K. Fan, X.~Qi, R.~Zhang, A.~Chen, C.~min Chan, W.~Xue, W.~Luo, S.~Zhang, and Y.~Guo, ``Eva: An embodied world model for future video anticipation,'' 2024. [Online]. Available: \url{https://arxiv.org/abs/2410.15461}
\BIBentrySTDinterwordspacing

\bibitem{asadi2019combatingcompoundingerrorproblemmultistep}
\BIBentryALTinterwordspacing
K.~Asadi, D.~Misra, S.~Kim, and M.~L. Littman, ``Combating the compounding-error problem with a multi-step model,'' 2019. [Online]. Available: \url{https://arxiv.org/abs/1905.13320}
\BIBentrySTDinterwordspacing

\bibitem{kalyanakrishnan2021analysisframeskippingreinforcementlearning}
\BIBentryALTinterwordspacing
S.~Kalyanakrishnan, S.~Aravindan, V.~Bagdawat, V.~Bhatt, H.~Goka, A.~Gupta, K.~Krishna, and V.~Piratla, ``An analysis of frame-skipping in reinforcement learning,'' 2021. [Online]. Available: \url{https://arxiv.org/abs/2102.03718}
\BIBentrySTDinterwordspacing

\bibitem{ding2024diffusionworldmodelfuture}
\BIBentryALTinterwordspacing
Z.~Ding, A.~Zhang, Y.~Tian, and Q.~Zheng, ``Diffusion world model: Future modeling beyond step-by-step rollout for offline reinforcement learning,'' 2024. [Online]. Available: \url{https://arxiv.org/abs/2402.03570}
\BIBentrySTDinterwordspacing

\bibitem{an2015variational}
J.~An and S.~Cho, ``Variational autoencoder based anomaly detection using reconstruction probability,'' \emph{Special lecture on IE}, vol.~2, no.~1, pp. 1--18, 2015.

\bibitem{nguyen2024variationalautoencoderanomalydetection}
\BIBentryALTinterwordspacing
H.~H. Nguyen, C.~N. Nguyen, X.~T. Dao, Q.~T. Duong, D.~P.~T. Kim, and M.-T. Pham, ``Variational autoencoder for anomaly detection: A comparative study,'' 2024. [Online]. Available: \url{https://arxiv.org/abs/2408.13561}
\BIBentrySTDinterwordspacing

\bibitem{9113298}
S.~Zavrak and M.~İskefiyeli, ``Anomaly-based intrusion detection from network flow features using variational autoencoder,'' \emph{IEEE Access}, vol.~8, pp. 108\,346--108\,358, 2020.

\bibitem{GAO2025112918}
\BIBentryALTinterwordspacing
Q.~Gao, C.~Liu, L.~Huang, G.~Trajcevski, Q.~Guo, and F.~Zhou, ``Learning to discover anomalous spatiotemporal trajectory via open-world state space model,'' \emph{Knowledge-Based Systems}, vol. 310, p. 112918, 2025. [Online]. Available: \url{https://www.sciencedirect.com/science/article/pii/S0950705124015521}
\BIBentrySTDinterwordspacing

\bibitem{sinha2024realtimeanomalydetectionreactive}
\BIBentryALTinterwordspacing
R.~Sinha, A.~Elhafsi, C.~Agia, M.~Foutter, E.~Schmerling, and M.~Pavone, ``Real-time anomaly detection and reactive planning with large language models,'' 2024. [Online]. Available: \url{https://arxiv.org/abs/2407.08735}
\BIBentrySTDinterwordspacing

\bibitem{huijben2024enhancingreconstructionbasedoutofdistributiondetection}
\BIBentryALTinterwordspacing
E.~M.~C. Huijben, S.~Amirrajab, and J.~P.~W. Pluim, ``Enhancing reconstruction-based out-of-distribution detection in brain mri with model and metric ensembles,'' 2024. [Online]. Available: \url{https://arxiv.org/abs/2412.17586}
\BIBentrySTDinterwordspacing

\bibitem{oquab2024dinov2learningrobustvisual}
\BIBentryALTinterwordspacing
M.~Oquab, T.~Darcet, T.~Moutakanni, H.~Vo, M.~Szafraniec, V.~Khalidov, P.~Fernandez, D.~Haziza, F.~Massa, A.~El-Nouby, M.~Assran, N.~Ballas, W.~Galuba, R.~Howes, P.-Y. Huang, S.-W. Li, I.~Misra, M.~Rabbat, V.~Sharma, G.~Synnaeve, H.~Xu, H.~Jegou, J.~Mairal, P.~Labatut, A.~Joulin, and P.~Bojanowski, ``Dinov2: Learning robust visual features without supervision,'' 2024. [Online]. Available: \url{https://arxiv.org/abs/2304.07193}
\BIBentrySTDinterwordspacing

\bibitem{dosovitskiy2021imageworth16x16words}
\BIBentryALTinterwordspacing
A.~Dosovitskiy, L.~Beyer, A.~Kolesnikov, D.~Weissenborn, X.~Zhai, T.~Unterthiner, M.~Dehghani, M.~Minderer, G.~Heigold, S.~Gelly, J.~Uszkoreit, and N.~Houlsby, ``An image is worth 16x16 words: Transformers for image recognition at scale,'' 2021. [Online]. Available: \url{https://arxiv.org/abs/2010.11929}
\BIBentrySTDinterwordspacing

\bibitem{Kingma_2019}
\BIBentryALTinterwordspacing
D.~P. Kingma and M.~Welling, ``An introduction to variational autoencoders,'' \emph{Foundations and Trends® in Machine Learning}, vol.~12, no.~4, p. 307–392, 2019. [Online]. Available: \url{http://dx.doi.org/10.1561/2200000056}
\BIBentrySTDinterwordspacing

\bibitem{NIPS2015_ced556cd}
T.~D. Kulkarni, W.~F. Whitney, P.~Kohli, and J.~Tenenbaum, ``Deep convolutional inverse graphics network,'' in \emph{Advances in Neural Information Processing Systems}, C.~Cortes, N.~Lawrence, D.~Lee, M.~Sugiyama, and R.~Garnett, Eds., vol.~28.\hskip 1em plus 0.5em minus 0.4em\relax Curran Associates, Inc., 2015.

\bibitem{5539957}
M.~D. Zeiler, D.~Krishnan, G.~W. Taylor, and R.~Fergus, ``Deconvolutional networks,'' in \emph{2010 IEEE Computer Society Conference on Computer Vision and Pattern Recognition}, 2010, pp. 2528--2535.

\bibitem{oord2018neuraldiscreterepresentationlearning}
\BIBentryALTinterwordspacing
A.~van~den Oord, O.~Vinyals, and K.~Kavukcuoglu, ``Neural discrete representation learning,'' 2018. [Online]. Available: \url{https://arxiv.org/abs/1711.00937}
\BIBentrySTDinterwordspacing

\end{thebibliography}

\end{document}